\documentclass[10pt,twocolumn,letterpaper]{article}

\usepackage[pagenumbers]{cvpr} % To force page numbers, e.g. for an arXiv version

\usepackage[pages=all]{background}

\backgroundsetup{
  scale=1,
  color=gray,
  opacity=0.6,
  angle=0,
  position={current page.south},
  vshift=1cm, % Distance from the bottom edge
  contents={\small Accepted to CVPR Findings 2026} % Replace with correct year
}

\definecolor{cvprblue}{rgb}{0.21,0.49,0.74}
\usepackage[pagebackref,breaklinks,colorlinks,allcolors=cvprblue]{hyperref}

\usepackage{makecell}
\usepackage{multirow}

\newcommand*\rot{\rotatebox{90}}

\def\confName{CVPR}
\def\confYear{2026}

\title{Complexity of Linear Regions in Self-supervised Deep ReLU Networks}

\author{Mufhumudzi Muthivhi, Terence L. van Zyl\\
University of Johannesburg, Johannesburg, South Africa\\
{\tt\small \{mmuthivhi, tvanzyl\}@uj.ac.za}
}

\begin{document}
\maketitle

\begin{abstract}
% ---ESTABLISHING THE CONTEXT---
There has been growing interest in studying the complexity of Rectified Linear Unit (ReLU) based activation networks. Recent work investigates the evolution of the number of piecewise-linear partitions (linear regions) that are formed during training. However, current research is limited to examining the complexity of models trained in a supervised way.
% ---FOCUS---
Self-Supervised Learning (SSL) differs in that it directly optimises the representation space using a loss function to enhance the model's performance across multiple downstream tasks. 
% ---STATING THE PURPOSE---
This study investigates the local distribution of linear regions produced by SSL models. We demonstrate that the evolution of linear regions correlates with the representation quality by utilising SplineCam to extract two-dimensional polytopes near the data distribution.
% ---DESCRIBING METHODOLOGY---
We track the number,  area, eccentricity, and boundaries of regions throughout training. The study compares supervised, contrastive, and self-distillation methods over two standard benchmark datasets, MNIST and FashionMNIST.
% ---PRESENTING THE RESULTS---
The analysis of the experimental results shows that self-supervised methods create substantially fewer regions to achieve comparable accuracy to supervised models. Contrastive methods rapidly expand regions over time, whereas self-distillation methods tend to consolidate by merging neighbouring regions. Lastly, we can detect representation collapse early within the geometric space of linear regions. Our analysis suggests that polytopal metrics can serve as reliable indicators of representation quality and model performance.
\end{abstract}    
\section{Introduction}
\label{sec:intro}

\begin{figure}[t]
  \centering
  \centerline{\includegraphics[width=\columnwidth]{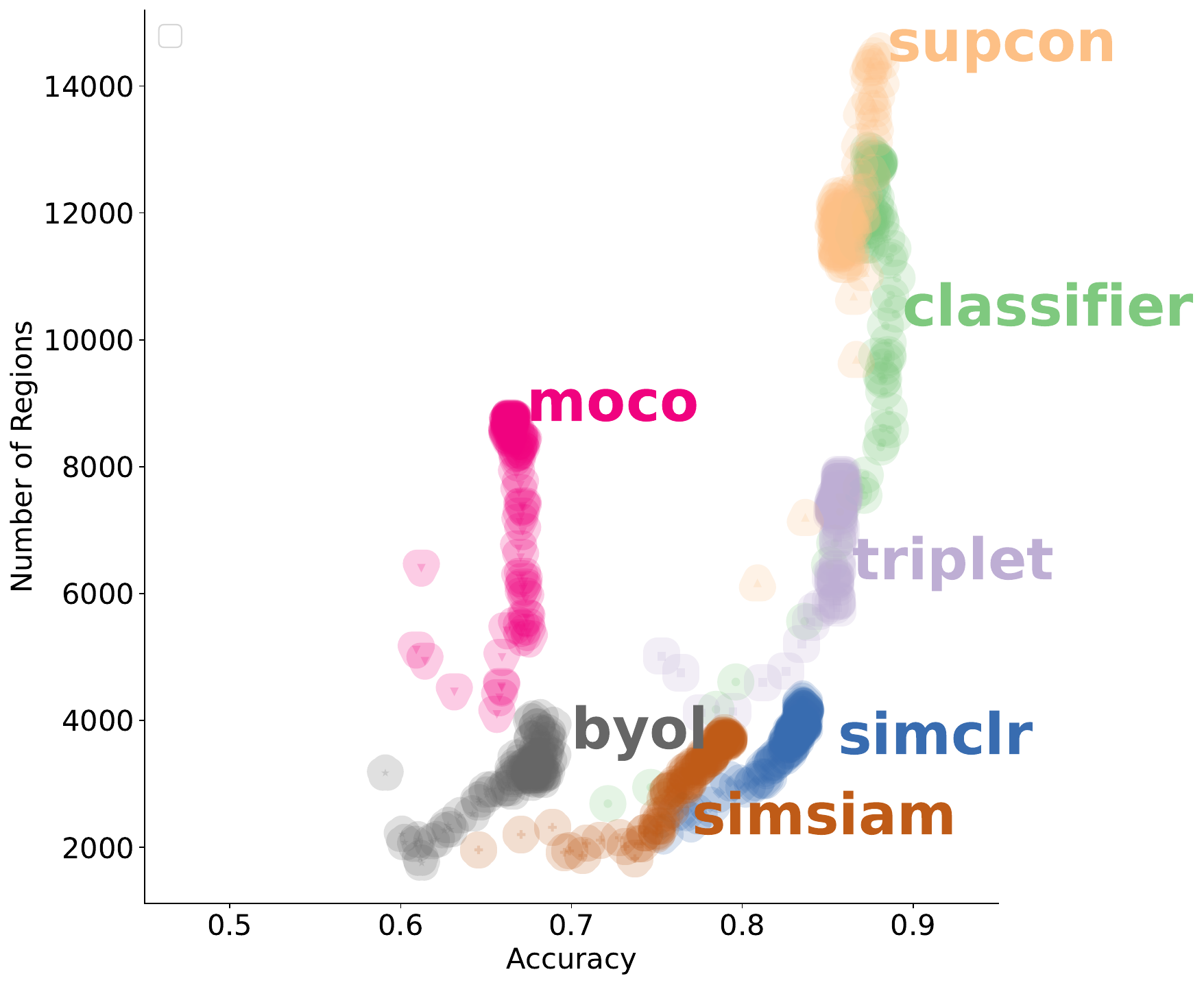}}

   \caption{Self-supervised methods (SimCLR, BYOL, SimSiam) produce fewer linear regions than their supervised counterparts (Classifier, Triplet, SupCon) while achieving comparable accuracy. Both contrastive and self-distillation approaches leverage positive pairings, resulting in the rapid merging of linear regions due to shared activation patterns. The opacity reflects the progression of training over epochs. Low-opacity points depict results achieved early in training, while darker points represent results achieved later.}
   \label{fig:regions_accuracy}
\end{figure}

% ---ESTABLISHING A RESEARCH TERRITORY---
Continuous Piecewise Linear (CPWL) neural networks partition the input space into a finite set of convex polytopes, referred to as \textit{linear regions}~\cite{hanin2019complexity}. Within each region, the network applies a single affine transformation. Regions are formed by applying a nonlinear activation function over the output values of the affine transformation. The set of all linear regions produced by a neural network defines a piecewise-linear tessellation of the input space. Understanding the structure and distribution of regions near the data distribution has become an active area of research. Practitioners count the number of regions as a measure of network complexity~\cite{montufar2014number, rolnick2017power, serra2018bounding}. \citeauthor{humayun2023splinecam} propose SplineCam for interpretation and visualisation of the exact number of regions from a two-dimensional input space~\cite{humayun2023splinecam}.
% ---ESTABLISHING A NICHE---
Most current work focuses on analysing the linear regions generated by supervised classifier networks~\cite{hanin2019complexity, humayun2024deep}. 
Some studies have explored the local geometry of generative models~\cite{humayun2022polarity, humayun2024secrets}. More recent work examines the complexity of input space partitioning in spiking neural networks~\cite{nguyen2025time}. 
However, the geometric properties of Self-Supervised Learning (SSL) representations remain largely underexplored. SSL has shown remarkable performance on several downstream tasks~\cite{ericsson2021well, caron2020unsupervised, simeoni2025dinov3}. It intrinsically generates supervision signals by augmenting input data.
% ---OCCUPYING THE NICHE---
% a) Outline your purposes and state the nature of your research
This work investigates the local distribution of linear regions produced by SSL methods. We transform a high-dimensional input space into a two-dimensional space and partition it to obtain its linear regions. Specifically, we consider contrastive and self-distillation methods~\cite{chen2020simple, he2020momentum, chen2021exploring, grill2020bootstrap}. Self-supervised contrastive learning captures the similarity among instances. They minimise the distance between positive pairs within the embedding space while simultaneously trying to keep negative pairs apart. However, they require large batch sizes or memory banks to store sufficient negative samples, thereby preventing overfitting. Self-distillation methods forgo the need for negatives by using a predictor network to predict the representation of views from the same image. Several specific conditions have been proposed to prevent representations from collapsing into degenerate solutions~\cite{tian2021understanding, jha2024common, zhang2022does}. 
% Some methods use a momentum encoder or a stop-gradient operation.  

% b) State your hypothesis or research question you seek to answer
We hypothesise that different SSL objectives produce unique geometric properties in the partitioned input space. In contrastive methods, positive pairs encourage samples to share activation patterns, essentially merging regions together. Negative pairs exert a repulsive force that tears regions apart. In contrast, self-distillation methods do not use negative pairs; thus, regions merge rapidly. This effect reflects on the heightened risk of representation collapse for self-distillation methods.
% c) Share your findings
\Cref{fig:regions_accuracy} contrasts the performance of supervised and self-supervised methods. All the SSL methods produce fewer regions and achieve accuracy comparable to that of supervised variants in terms of performance. Furthermore, results show that the contrastive objective produces a larger number of isotropic small regions that fit the in-distribution data well. Self-distillation encourages fewer, larger, and more anisotropic partitions, which correlates with improved generalisation performance.
% d) Elaborate on the value of your research
We contribute to the existing literature by:
\begin{itemize}
    \item track the number, area, eccentricity, and boundaries of regions throughout training from a projected high-dimensional input space,
    \item compare the evolution of regions from supervised, contrastive and self-distillation methods,
    \item simulates the effects of representation collapse within the polytopal space,
    \item provides a qualitative analysis of linear regions and their density within the projected high-dimensional input space. 
\end{itemize}

\section{Related Work}
\label{sec:relatedwork}

\subsection{Continuous Piecewise-Linear Functions}

A neural network consists of one or more linear transformations followed by a non-linear operation. Current deep networks predominantly use a Continuous Piecewise-Linear (CPWL) nonlinearity such as a Rectified Linear Unit (ReLU), leaky-ReLU or MaxOut operation~\cite{glorot2011deep}. CPWL functions allow practitioners to assess the complexity of a neural network by counting the number of linear partitions (linear regions) formed by a network. \citeauthor{montufar2014number} show that the number of linear regions grows exponentially with depth~\cite{montufar2014number}. \citeauthor{raghu2017expressive} relate the number of regions to expressivity and generalization~\cite{raghu2017expressive}. \citeauthor{novak2018sensitivity} demonstrate that neural networks have strong robustness near the training data distribution, especially with data augmentation and ReLU activations~\cite{novak2018sensitivity}. \citeauthor{serra2018bounding} provides tighter bounds on the number of regions compared to previous studies~\cite{serra2018bounding}. Several works utilise CPWL functions to investigate properties in generative models, Recurrent Neural Networks, or spiking neural networks, including the understanding of classifiers, out-of-distribution detection, and the interpretation and visualisation of regions emerging from supervised classifiers~\cite{humayun2022polarity, humayun2024deep, casco2024visualizing, gamba2022all, ji2022test, nguyen2025time, humayun2023splinecam, hanin2019complexity}.

\subsection{Self-supervised Learning}

Recently, Self-supervised Learning (SSL) methods have achieved performance comparable to that of their supervised counterparts across a range of downstream tasks~\cite{simeoni2025dinov3}. SSL utilises a formulated pretext-task to generate several pseudo targets from the input data~\cite{gui2024survey}. An objective function learns a representation space that is invariant to augmentations made to the input data. Early work utilised contrastive learning to associate positive pairs with each other while pushing negative pairs apart~\cite{chen2020simple, he2020momentum}. Self-distillation methods maximise the similarity between two augmented views of the same image~\cite{chen2021exploring, grill2020bootstrap}. One limitation is that they do not explicitly include a repulsive force (negative examples) to prevent the representations from collapsing into degenerate solutions. Hence, the majority of work in this area focuses on devising implicit architectural constraints and optimisation strategies, such as a prediction head, stop-gradient operation, or a redundancy-reduction objective~\cite{zbontar2021barlow, caron2021emerging, caron2020unsupervised}.

\begin{figure*}[t]
  \centering
  \centerline{\includegraphics[width=\textwidth]{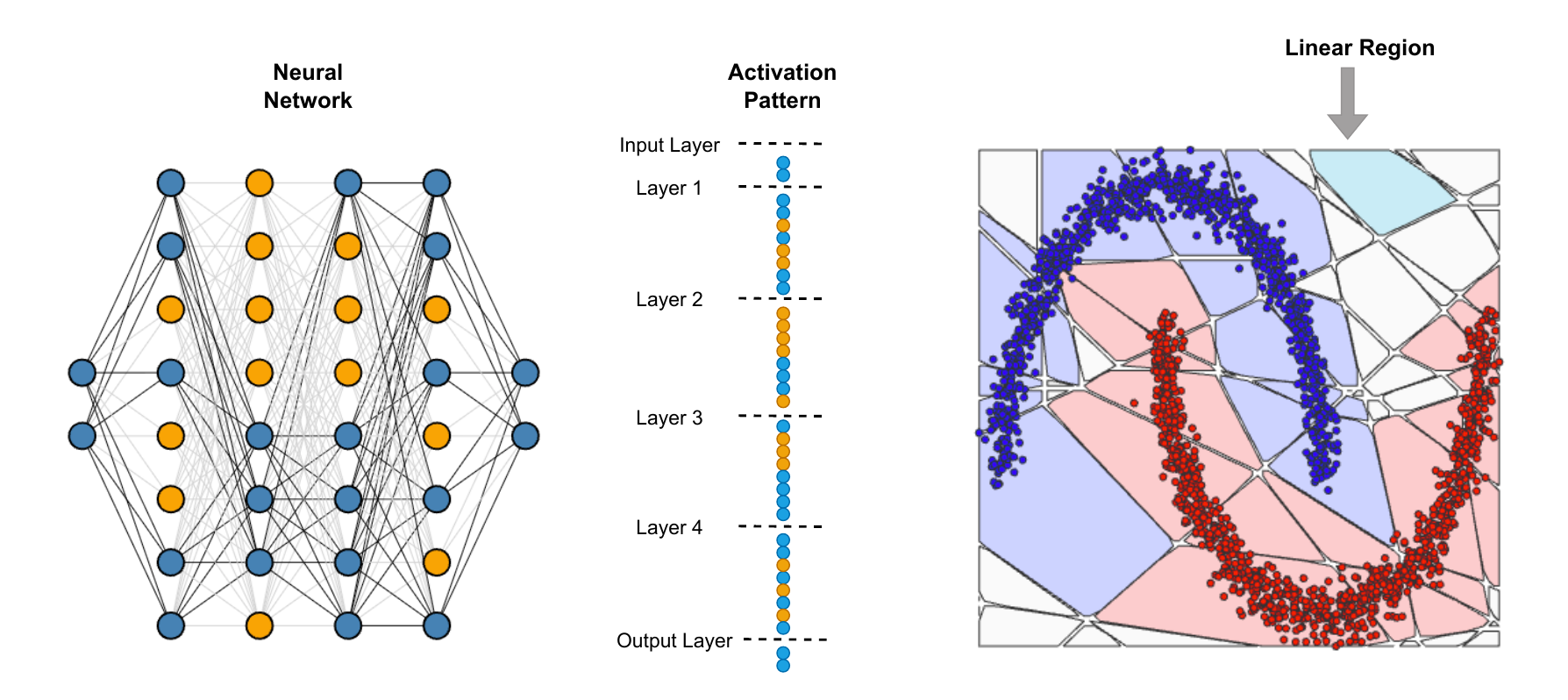}}

   \caption{Illustration of how a ReLU network partitions the input space into linear regions. The orange neurons are active. The distinct activation pattern corresponds to a single linear region highlighted in blue on the right plot. Each region applies a unique affine transformation to the partitioned values.}
   \label{fig:linear_region}
\end{figure*}

\subsection{Representation Learning}

Self-supervised representations have been shown to convey stronger transferability and generalisation characteristics~\cite{ericsson2021well, caron2020unsupervised, dhamija2021self, simeoni2025dinov3}. Practitioners assess the quality of representations by freezing the encoder and extracting a feature vector for
each image~\cite{wu2018unsupervised}. \citeauthor{garrido2023rankme} propose to analyse the singular value distribution of the learned embeddings to assess representation quality~\cite{garrido2023rankme}. \citet{chen2021exploring} uses the mean of the standard deviation of the L2-normalised feature vector to determine the level of representation collapse. \citet{jha2024common} suggests that minimising the magnitude of the expected representation over all data samples while simultaneously maximising the magnitude of individual samples over different data augmentations helps prevent representation collapse. This work extends these efforts by introducing a polytopal analysis of learned representations. By quantifying the number, geometric shape, and evolution of linear regions within the network’s activation space, we provide a geometric view of the encoder's features across supervised, contrastive, and self-distillation methods.

\section{Methodology}

\subsection{Definitions}

Let $f$ denote a Continuous Piecewise Linear (CPWL) neural network composed of layers with nonlinear activation functions $\sigma$. For an input $x \in \mathbb{R}^d$ and parameters $\theta = \{W, b\}$, a single-layer transformation can be written as
\begin{equation}
    f(x; \theta) = \sigma(Wx + b),
\end{equation}
where $W \in \mathbb{R}^{m \times d}$ and $b \in \mathbb{R}^m$ represent the weight matrix and bias vector, respectively.
For ReLU-based networks, the activation function is defined as
\begin{equation}
    \sigma(z) = \max(0, z),
\end{equation}
which creates a partitioning of the input space $\mathbb{R}^d$ into a finite set of convex polytopes~\cite{glorot2011deep}. Each region corresponds to a unique activation pattern of the hidden units, such that within each region, $f$ behaves as an affine transformation. Figure~\ref{fig:linear_region} depicts a unique activation pattern produced by a neural network and its corresponding region mapped to a two-dimensional input grid. Formally, the input space can be expressed as a union of linear regions:
\begin{equation}
    \mathbb{R}^d \approx \bigcup_{i=1}^{N_\mathcal{R}} \mathcal{R}_i,
\end{equation}
where $\mathcal{R}_i$ denotes the $i$-th region and $N_\mathcal{R}$ is the \textit{total number of distinct regions} formed by the activation boundaries. The number and structure of these regions serve as a measure of the network’s complexity~\cite{humayun2023splinecam, hanin2019complexity}.

\begin{figure}[t]
  \centering
  \centerline{\includegraphics[width=\columnwidth]{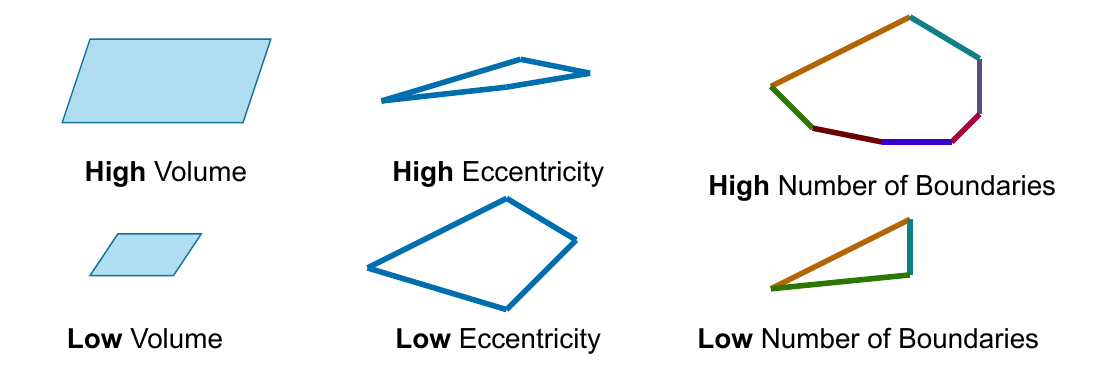}}

   \caption{Illustration of the geometric properties of linear regions. Each region is characterised by its volume, eccentricity, and boundary count.}
   \label{fig:geometry}
\end{figure}

\subsection{Geometric Properties}

We consider the \textit{volume}, \textit{eccentricity}, and \textit{boundaries} of each polytope to characterise the geometric properties of the learned representation space~\cite{hanin2019complexity}. Volume quantifies the extent to which the input space is subjected to a single affine transformation. A region with a larger volume suggests a greater contribution to the final representation of $x$. Eccentricity captures the degree of anisotropy. Circular regions have a lower eccentricity, ensuring that the learnt representations remain farther from activation boundaries. Elongated regions with high eccentricity are often located near the activation boundaries. Small changes in input can trigger shifts in the active set of neurons. Lastly, we refer to the vertices of the polytope as the region's boundaries. More boundaries suggest a complex partitioning of the input space. Highly partitioned input space could lead to less stability since a perturbation of the data point $x$ can cause it to jump between several regions. Figure~\ref{fig:geometry} provides an illustration of the effects of each geometric property on data point $x$. 

\paragraph{Volume:}
In two dimensions, we compute the volume of a region $\mathcal{R}_i$ with vertices $\{(a_0, b_0), \ldots, (a_{n-1}, b_{n-1})\}$ using the shoelace formula:
\begin{equation}
    V_i = \frac{1}{2}\left|\sum_{k=0}^{n-1}\big(a_k b_{k+1} - b_k a_{k+1}\big)\right|,
    \label{eq:area}
\end{equation}
such that the last set of points is equal to the first $(a_n, b_n) = (a_0, b_0)$.

\paragraph{Eccentricity:}
To characterise the anisotropy of each region, we compute an eccentricity measure based on its axis-aligned bounding box. Let
\begin{equation}
    \Delta_x = \max(a_k) - \min(a_k), \qquad 
    \Delta_y = \max(b_k) - \min(b_k),
\end{equation}
given the major $\max(\Delta_a, \Delta_b)$ and minor $\min(\Delta_a, \Delta_b)$ axis lengths, we describe the eccentricity as
\begin{equation}
    E_i = \sqrt{1 - \left(\frac{\min(\Delta_a, \Delta_b)}{\max(\Delta_a, \Delta_b)}\right)^2},
    \label{eq:eccentricity}
\end{equation}
with $E_i = 0$ for degenerate (single-point) regions. 

\paragraph{Boundaries:}

The number of boundaries $B_i=n$ is simply the total number of unique edges for a two-dimensional polytope.

\begin{figure}
  \centering
  \centerline{\includegraphics[width=\columnwidth]{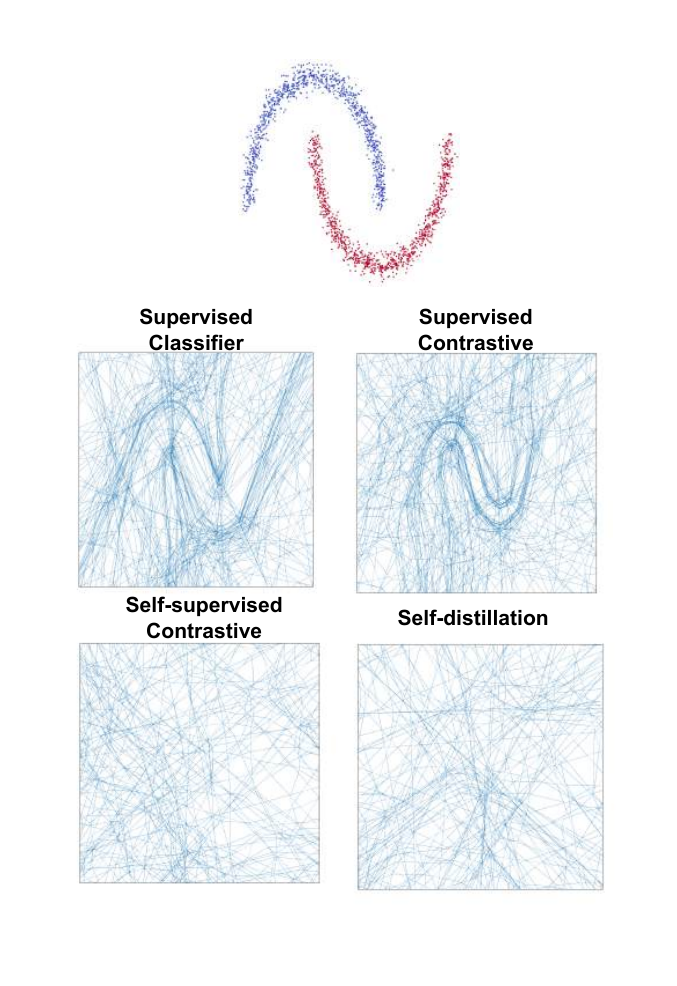}}

   \caption{Density of linear regions produced near the moon's data distribution (left). The regions are generated by a backbone neural network with a width 64 and a depth of 5.}
   \label{fig:moons_plots}
\end{figure}

\begin{table}[t]
  \centering
  \caption{Number and properties of regions generated by a backbone neural network with a width 64 and depth of 5 trained over the moons dataset.}
  \label{tab:moons_result}
  \resizebox{\columnwidth}{!}{%
  \begin{tabular}{lcccccc}
    \toprule
    \textbf{Model} & \makecell[l]{Number Of\\Regions} & \makecell[l]{Mean\\Area ($\times 10^{-4}$)} & \makecell[l]{Mean\\Eccentricity} & \makecell[l]{Mean\\Boundaries}  \\
    \midrule
    \makecell[l]{Supervised-\\Classifier}       & 9232 & 6.70 & 0.793 & 3.52 \\
    \makecell[l]{Supervised-\\Contrastive}      & 9679 & 6.42 & 0.774 & 3.52 \\
    \makecell[l]{Self-supervised-\\Contrastive} & 4354 & 13.8 & 0.717 & 3.53 \\
    Self-distillation           & 3602 & 15.5 & 0.705 & 3.55 \\
    \bottomrule
  \end{tabular}
  }
\end{table}

\subsection{2-dimensional Input Space}

We first provide an intuitive visualisation of our framework by using a two-dimensional input space that sits near the data distribution. To examine how the network partitions the input space, we define a uniform sampling grid
\begin{equation}
    D = \{(a_k, b_k) \mid a_{k}, b_{k} \in [0,1] \}.
\end{equation}
Each point $(a_k, b_k)$ is propagated through the network to record the activation pattern across all ReLU operations. 
The subset of points in $D$ that produce the same activation pattern belongs to the same linear region $\mathcal{R}_i$. Thus, the input domain can be approximated as
\begin{equation}
    D \approx \bigcup_{i=1}^{N_{\mathcal{R}}} \mathcal{R}_i,
\end{equation}
which is the union of linear regions. By dividing the mesh according to the activation patterns, we obtain a piecewise-linear tessellation of the input space, as illustrated in Figure~\ref{fig:moons_plots}. 
We train a neural network $f$ on the two-dimensional Moons dataset and visualise how the model develops linear regions that partition the input domain. \Cref{tab:moons_result} presents the number of regions and geometric properties of the linear regions produced by a supervised classifier, supervised contrastive, a self-supervised contrastive and a self-distillation model. The self-distillation model has the largest volume of linear regions on average, and the classifier exhibits the highest eccentricity and the largest number of regions. This result correlates visually with the illustration in Figure~\ref{fig:moons_plots}  

\begin{figure*}[htp]
  \centering
  \begin{subfigure}[t]{0.48\textwidth}
    \centering
    \includegraphics[width=\textwidth]{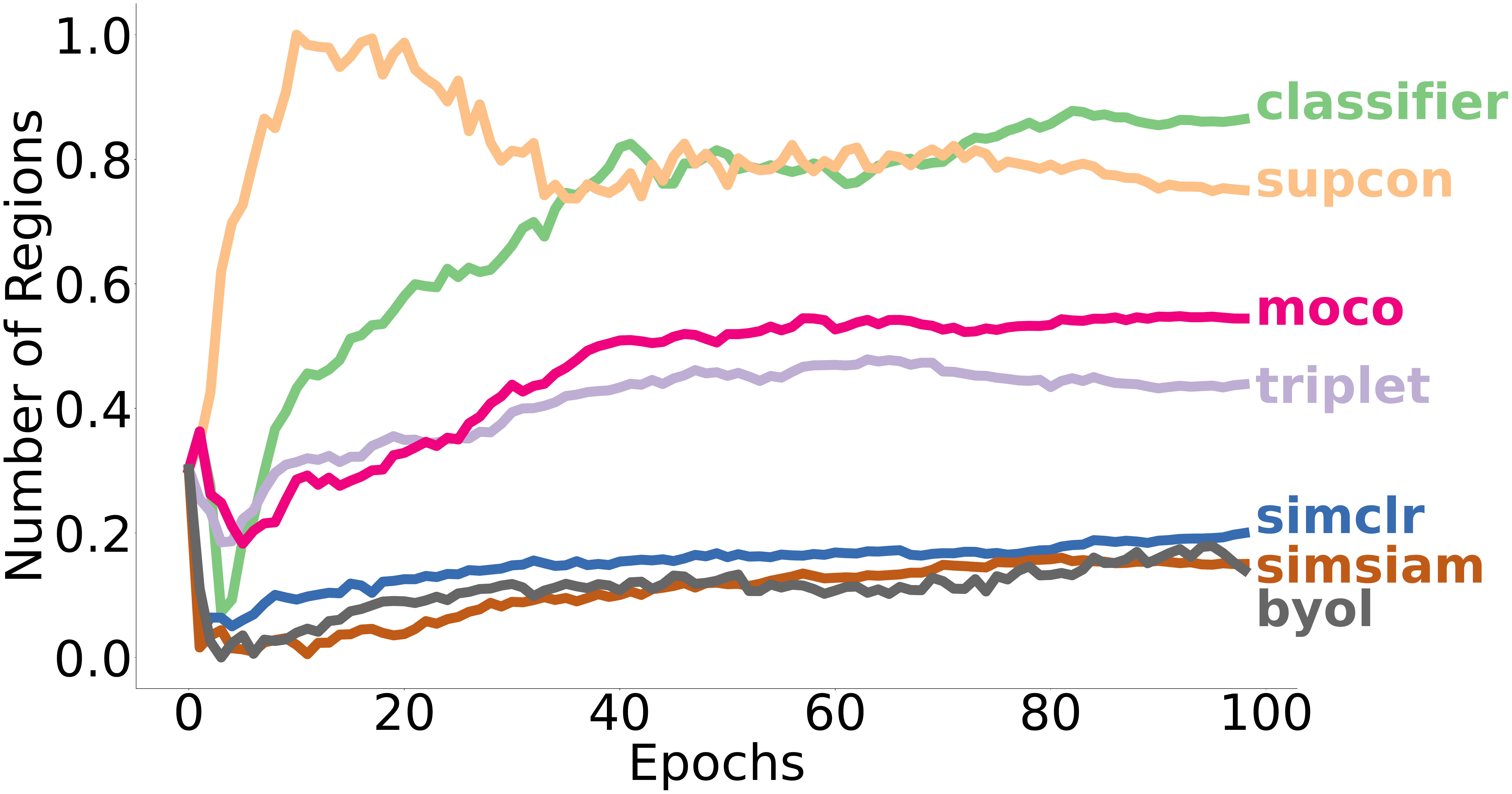}
    \caption{MNIST}
    \label{fig:num_regions_mnist}
  \end{subfigure}
  \hfill
  \begin{subfigure}[t]{0.48\textwidth}
    \centering
    \includegraphics[width=\textwidth]{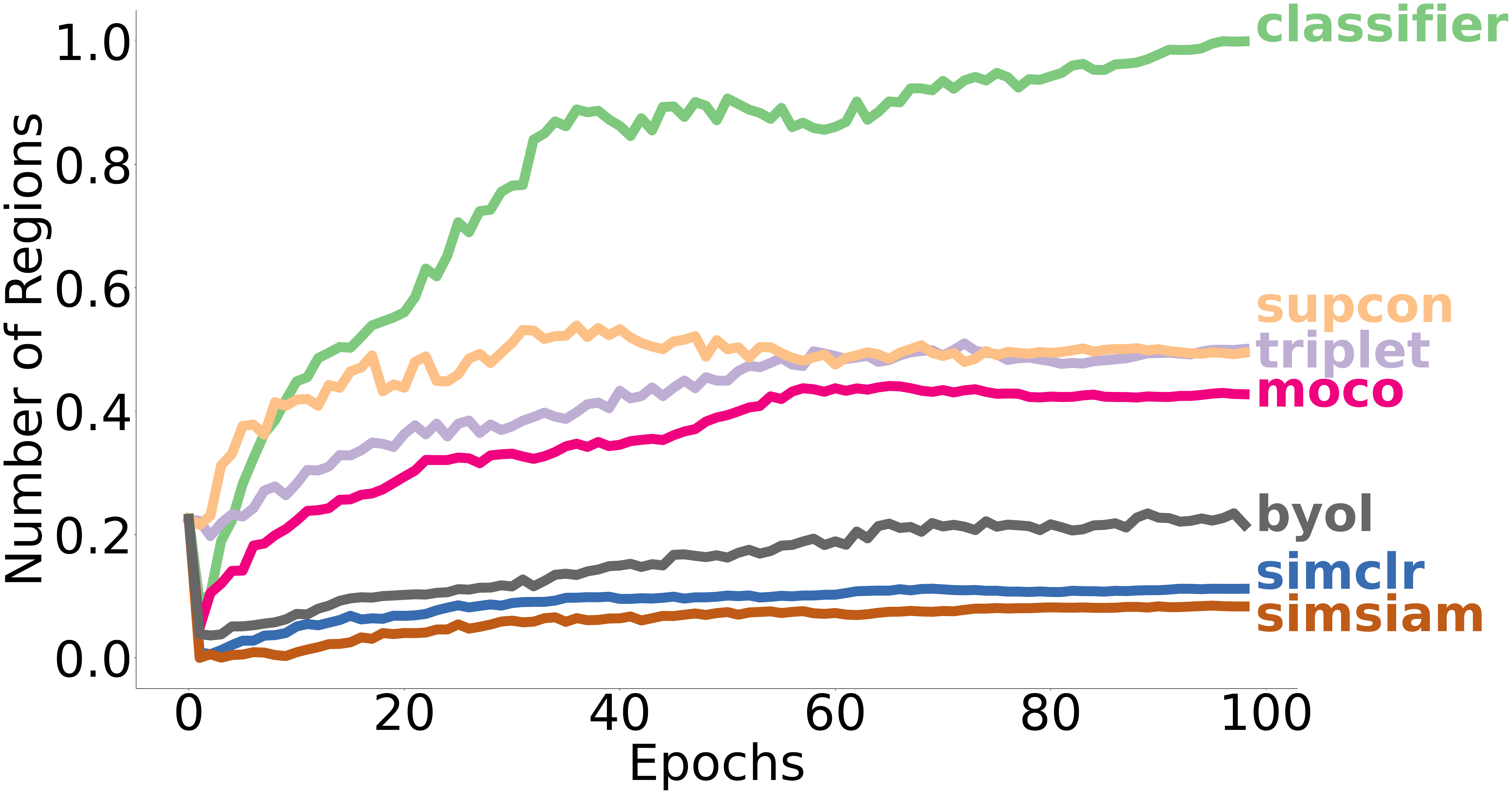}
    \caption{FashionMNIST}
    \label{fig:num_regions_fashionmnist}
  \end{subfigure}
  \caption{Evolution of linear regions over 100 epochs generated by a backbone encoder. The supervised methods (Classifier, SupCon and MoCo) observe a slight decrease in the number of regions, followed by a rapid increase as training progresses. In contrast, self-supervised methods experience an early decline before gradually increasing. On average, contrastive self-supervised methods (SimCLR and MoCo) expand their region counts more rapidly than the self-distillation methods (SimSiam and BYOL). The use of negative samples in contrastive methods promotes partitioning of the input space over time. The absence of a repulsive force in self-distillation approaches results in slower growth and even stagnation of region complexity.}\label{fig:num_regions}  
\end{figure*}

\subsection{High-dimensional Input Space}

For a higher-dimensional input space, we adopt a projection-based strategy to partition the input space. Specifically, following the experimental setup of \citeauthor{novak2018sensitivity} and \citeauthor{hanin2019complexity}, we visualise the local structure of linear regions by projecting a small neighbourhood of the data distribution onto a two-dimensional plane within the $d$-dimensional input space~\cite{novak2018sensitivity, hanin2019complexity}. 
The plane is defined by three data points from distinct classes (zero, one, and two) and centred at the circumcenter of these examples. We employ \textit{SplineCam} to automatically extract and visualise the continuous piecewise-linear partitions~\cite{humayun2023splinecam}. SplineCam utilises spline-based interpolation to map activation boundaries.

\section{Experimental Setup}

\begin{table}[t]
  \centering
  \caption{Total number of parameters (in thousands) for the encoder, projection, and prediction heads across datasets.}
  \label{tab:number_of_params}
  \resizebox{\columnwidth}{!}{%
  \begin{tabular}{lcccccc}
    \toprule
    \textbf{Model} & 
    \multicolumn{2}{c}{\textbf{Backbone}} & 
    \multicolumn{2}{c}{\textbf{Projection}} & 
    \multicolumn{2}{c}{\textbf{Prediction}} \\
    \cmidrule(lr){2-3} \cmidrule(lr){4-5} \cmidrule(lr){6-7}
    & \textbf{MNIST} & \textbf{\makecell[l]{Fashion-\\MNIST}} 
    & \textbf{MNIST} & \textbf{\makecell[l]{Fashion-\\MNIST}} 
    & \textbf{MNIST} & \textbf{\makecell[l]{Fashion-\\MNIST}} \\
    \midrule
    \textbf{Supervised} \\
    \quad Classifier    & \multirow{10}{*}{\rot{$66.9$ for each}} & \multirow{10}{*}{\rot{$166$ for each}} & 0.6 & 18 & – & – \\
    \quad Triplet Loss  &&& – & – & – & – \\
    \quad SupCon        &&& - & - & – & – \\
    \textbf{Self-Supervised} \\
        \quad \textit{Contrastive} \\
        \quad \quad SimCLR        &&& 6.3 & 25 & – & - \\
        \quad \quad MoCo          &&& 6.2 & 24.8 & – & - \\
        \quad \textit{Self-distillation} \\
            \quad \quad SimSiam       &&& 10.5 & 41.5 & 1.5 & 4.2 \\
            \quad \quad BYOL          &&& 12.6 & 49.7 & 4.3 & 6.7 \\
    \bottomrule
  \end{tabular}
  }
\end{table}

\subsection{Models}

For each supervised and self-supervised variant, we use the same encoder architecture depending on the dataset used. \Cref{tab:number_of_params} presents the number of parameters for each method. We adopt the same scaling factors for the projection and prediction heads as specified in the original proposed implementations. We investigate three supervised methods.
The \textbf{classifier} uses a linear layer to compute a linear combination of the activations from the previous encoder layer~\cite{pascanu2013number}. \textbf{Triplet} loss associates an image with another positive image belonging to the same class while pushing the negative image away~\cite{hermans2017defense}. Supervised Contrastive (\textbf{SupCon}) learning extends contrastive learning by pulling together representations of samples from the same class while treating samples from different classes within the mini-batch as negatives~\cite{khosla2020supervised}.
We also study three more self-supervised methods. Simple Framework for Contrastive Learning (\textbf{SimCLR}) is also a contrastive learning objective that does not use labels~\cite{chen2020simple}. SimCLR applies augmentations to a single image to derive positive pairs and samples negatives from the mini-batch. Momentum Contrast (\textbf{MoCo}) uses a queue of negative samples generated by a momentum encoder~\cite{he2020momentum}. Similarity Siamese (\textbf{SimSiam}) and Bootstrap Your Own Latent (\textbf{BYOL}) are both self-distillation methods that do not make use of negative samples~\cite{chen2021exploring, grill2020bootstrap}. SimSiam uses a dual parallel encoder with a prediction head and a stop gradient operation to prevent the model from collapsing. BYOL further applies two separate identical encoders. A frozen encoder has a moving average over the weights to update the target encoder. 

\subsection{Datasets}

Following prior work, we train and present results over two benchmark datasets~\cite{humayun2023splinecam, hanin2019complexity, hanin2019deep}. MNIST consists of 70,000 grayscale images of handwritten digits~\cite{lecun2002gradient}. FashionMNIST is a more challenging alternative containing images of clothing items~\cite{xiao2017fashion}. All images are grayscale with 60,000 samples for training and 10,000 for testing.

\subsection{Implementation Details}

The supervised loss functions are adopted from PyTorch Metric Learning, and the self-supervised models are taken from Lightly~\cite{musgrave2020pytorch, lightly2025github}. Each model runs for 100 epochs. We apply random resized cropping, horizontal flipping, greyscaling and Gaussian blur to our augmentation strategy. All training configurations and settings are listed in the supplementary material. We use a single NVIDIA GeForce GTX 5070 Ti GPU to accelerate training.
\section{Results}

\begin{table}[htbp]
\caption{Accuracy and linear regions properties over the MNIST dataset.}
\begin{center}
\resizebox{\columnwidth}{!}{%
\begin{tabular}{lrrrrrrrrrr}
\toprule
Models  \\
& Accuracy & Regions & Volume & Eccentricity & Boundaries  \\
\toprule
\multicolumn{3}{l}{\textbf{\textit{supervised contrastive}}} \\
\; Classifier                  & 0.8800 & 12817 & 0.1164 & 0.7521 & 3.534 \\
\; Triplet loss                & 0.8531 & 7365 & 0.1789 & 0.7563 & 3.523 \\
\; SupCon              & 0.8557 & 11349 & 0.1042 & 0.7435 & 3.549 \\
\bottomrule
\multicolumn{3}{l}{\textbf{\textit{self-supervised contrastive}}} \\
\; SimCLR                      & 0.8351 & 4297 & 0.3362 & 0.7632 & 3.526  \\
\; MoCo                        & 0.6641 & 8731 & 0.1644 & 0.7698 & 3.544\\
\textbf{\textit{self-distilliation}} \\
\; Simsiam                     & 0.7915 & 3653 & 0.3986 & 0.7725 & 3.567 \\
\; BYOL                        & 0.6772 & 3730 & 0.3924 & 0.7989 & 3.539 \\
\bottomrule
\end{tabular}
}
\label{table:results_mnist}
\end{center}
\end{table}

\begin{table}[htbp]
\caption{Accuracy and linear regions properties over the FashionMNIST dataset.}
\begin{center}
\resizebox{\columnwidth}{!}{%
\begin{tabular}{lrrrrrrrrrr}
\toprule
Models  \\
& Accuracy & Regions & Volume & Eccentricity & Boundaries  \\
\toprule
\multicolumn{3}{l}{\textbf{\textit{supervised contrastive}}} \\
\; Classifier                  & 0.8012 & 78178 & 0.0235 & 0.7339 & 3.549 \\
\; Triplet loss                & 0.8204 & 41413 & 0.0413 & 0.7386 & 3.547 \\
\; SupCon              & 0.8072 & 40988 & 0.0371 & 0.7440 & 3.542 \\
\bottomrule
\multicolumn{3}{l}{\textbf{\textit{self-supervised contrastive}}} \\
\; SimCLR                      & 0.7568 & 8080 & 0.2139 & 0.7378 & 3.542  \\
\; MoCo                        & 0.7091 & 35996 & 0.0475 & 0.7399 & 3.550 \\
\multicolumn{3}{l}{\textbf{\textit{self-distilliation}}} \\
\; Simsiam                     & 0.7523 & 10731 & 0.1602 & 0.7430 & 3.549 \\
\; BYOL                        & 0.7644 & 20946 & 0.0898 & 0.7280 & 3.555 \\
\bottomrule
\end{tabular}
}
\label{table:results_fashionmnist}
\end{center}
\end{table}

\subsection{Number Of Regions}

\Cref{fig:num_regions} depicts the evolution of linear regions of each model from the first to the 100th epoch. Supervised and self-supervised learning methods exhibit distinct learning progressions. The supervised methods exhibit a rapid rise in the number of regions. Within the first 10 epochs, SupCon regions increase 137\% and 63\% for the MNIST and FashionMNIST datasets, respectively. Around the 20th epoch, a slight and steady decrease in regions follows. An early plateau suggests a merging of regions to accurately fit the data distribution.
The self-supervised methods experience an inverse relationship. The number of regions after initialisation plateaued at 47\%, 62\%, and 62\% for SimCLR, SimSiam, and BYOL, respectively, on the MNIST dataset. A similar relationship is observed for the FashionMNIST dataset. The MoCo and SimCLR follow a smoother trajectory of steady region growth throughout training. Simsiam and BYOL increase regions slightly, then slowly merge regions throughout training. 

\subsection{Accuracy and Regions}

\Cref{table:results_mnist} and \Cref{table:results_fashionmnist} present the number of regions produced by the encoder and its accuracy on the validation set over the MNIST and FashionMNIST datasets, respectively. Supervised methods require high region counts to achieve a moderately high accuracy rate. SupCon requires 7,200 regions to achieve an accuracy of 83.69\%, whereas SimCLR requires only 4,297 regions at an accuracy of 83.53\%. Although the SSL methods converge at lower accuracy, they require substantially fewer regions to achieve comparable accuracy. The reduction in complexity is a desirable property, as prior work shows that models with fewer regions exhibit improved robustness and generalization~\cite{croce2019provable}. Moreover, \citeauthor{humayun2024deep} demonstrates that the delayed generalisation phenomenon, known as grokking, also emerges at reduced complexity~\cite{humayun2024deep}. SSL methods produce an almost linear relationship between the number of regions and to accuracy achieved. After merging regions during initialisation, simply splitting the same regions correctly aligns with the data distribution. 

\begin{figure*}[t]
  \centering
  \centerline{\includegraphics[width=\textwidth]{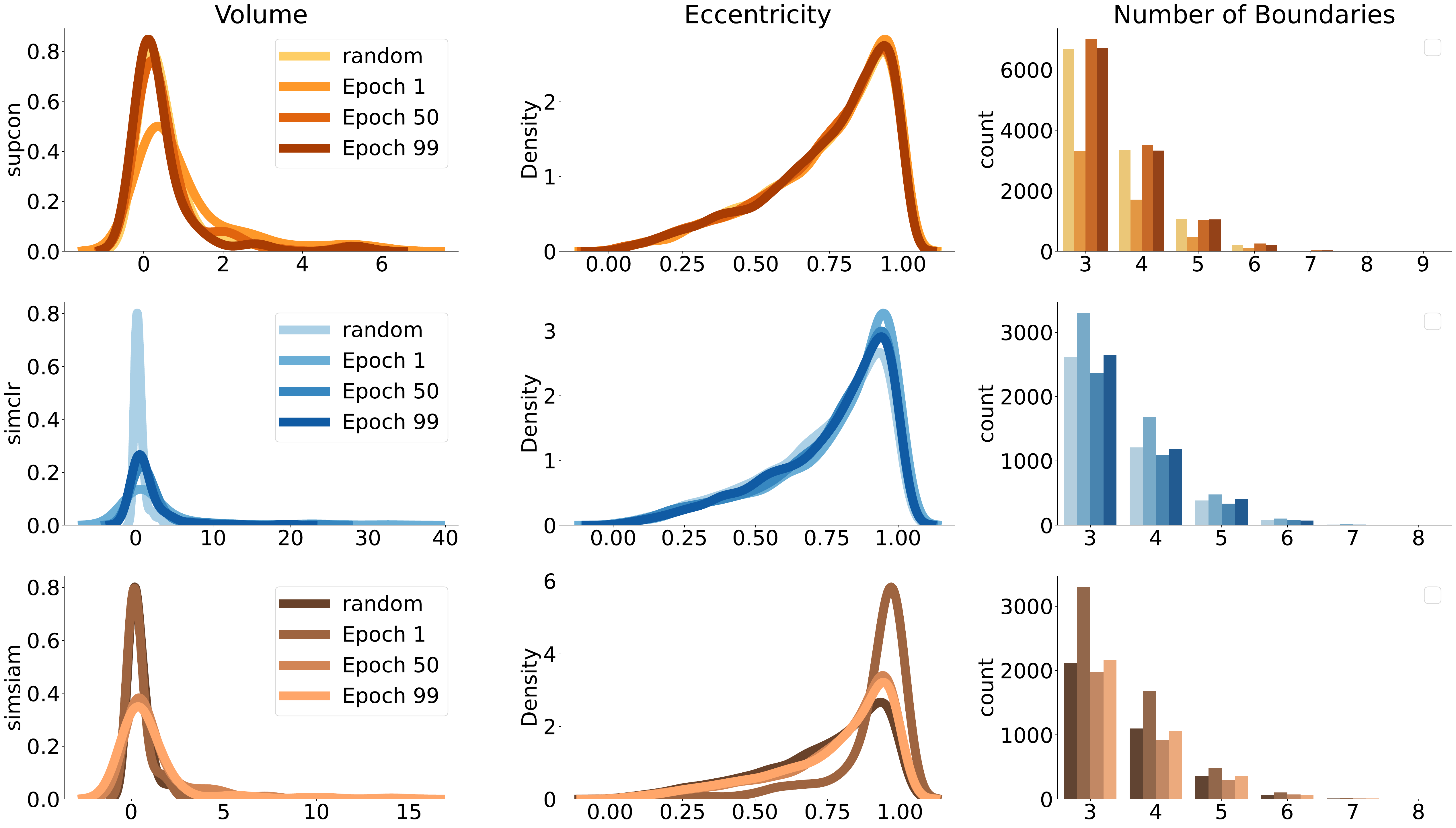}}

   \caption{Evolution of region area, eccentricity, and boundary count distributions across epochs for SupCon, SimCLR, and SimSiam on MNIST dataset.}
   \label{fig:distribution}
\end{figure*}

\subsection{Geometry Of Regions}

The SSL methods produce regions that are, on average, about 2.42 and 3.76 times the volume of the supervised methods for MNIST and FashionMNIST, respectively. The final mean eccentricity and number of boundaries are around less than 0.8 and 3.6 across each of the models. \Cref{fig:distribution} illustrates the evolution of the three geometric properties from initialisation to the 1st, 50th, and 99th epoch on the MNIST dataset. The eccentricity and number of boundaries exhibit somewhat skewed distributions, suggesting that the mean may not provide a fully representative characterisation. However, SimCLR and SimSiam slightly alter the tails of their eccentricity distributions. This progression indicates that a subset of the regions becomes more elongated as training progresses. The self-distillation methods produce, on average, a higher final mean eccentricity than the SSL contrastive methods. 

\subsection{Representation Collapse}

\begin{figure}[t]
  \centering
  \centerline{\includegraphics[width=\columnwidth]{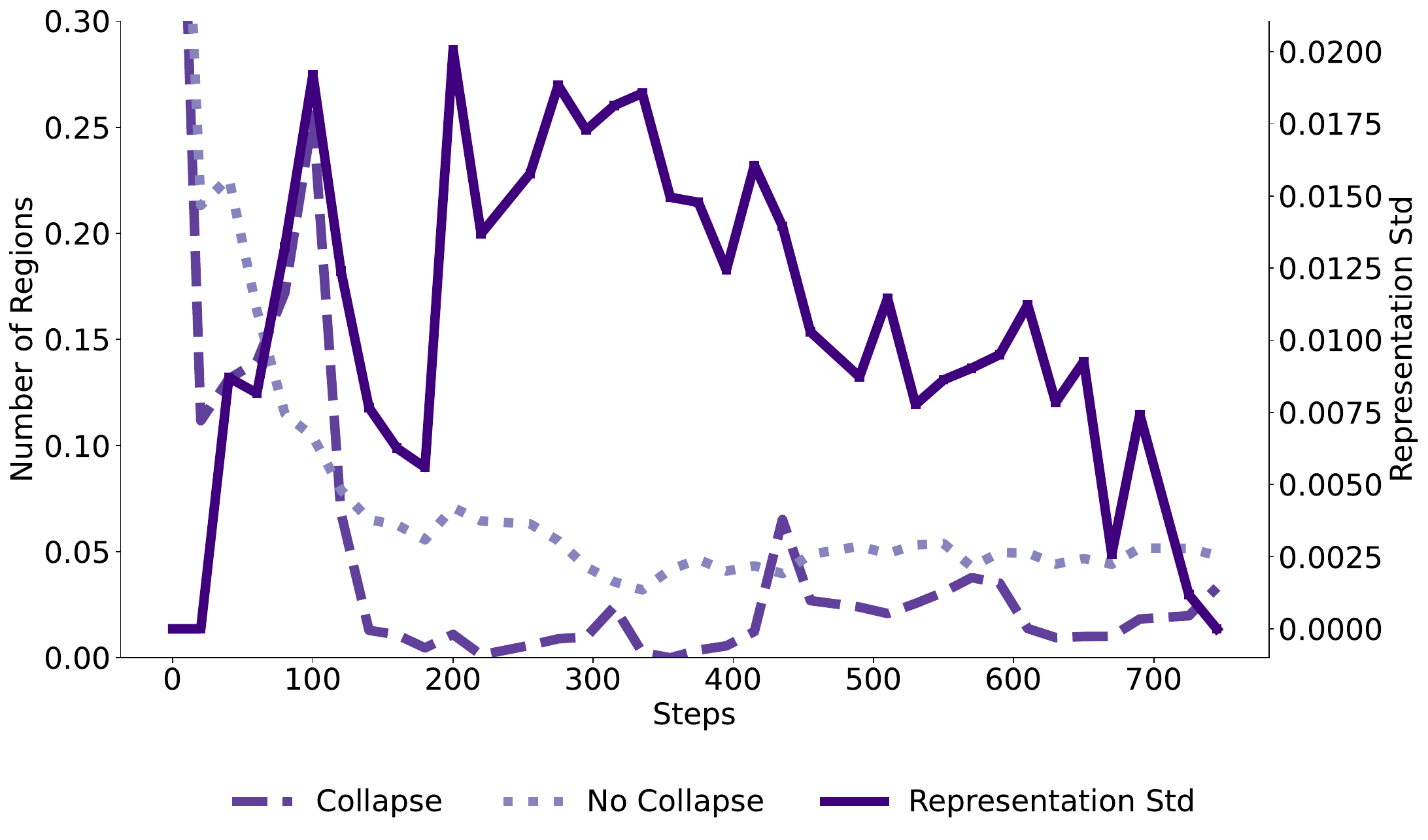}}

   \caption{Relationship between the number of linear regions and representation standard deviation during training for Simsiam without a prediction head to simulate collapse. The number of regions decreases rapidly earlier in training compared to the standard deviation of the representation.}
   \label{fig:num_regions_rep_std}
\end{figure}

\begin{figure}[t]
  \centering
  \centerline{\includegraphics[width=\columnwidth]{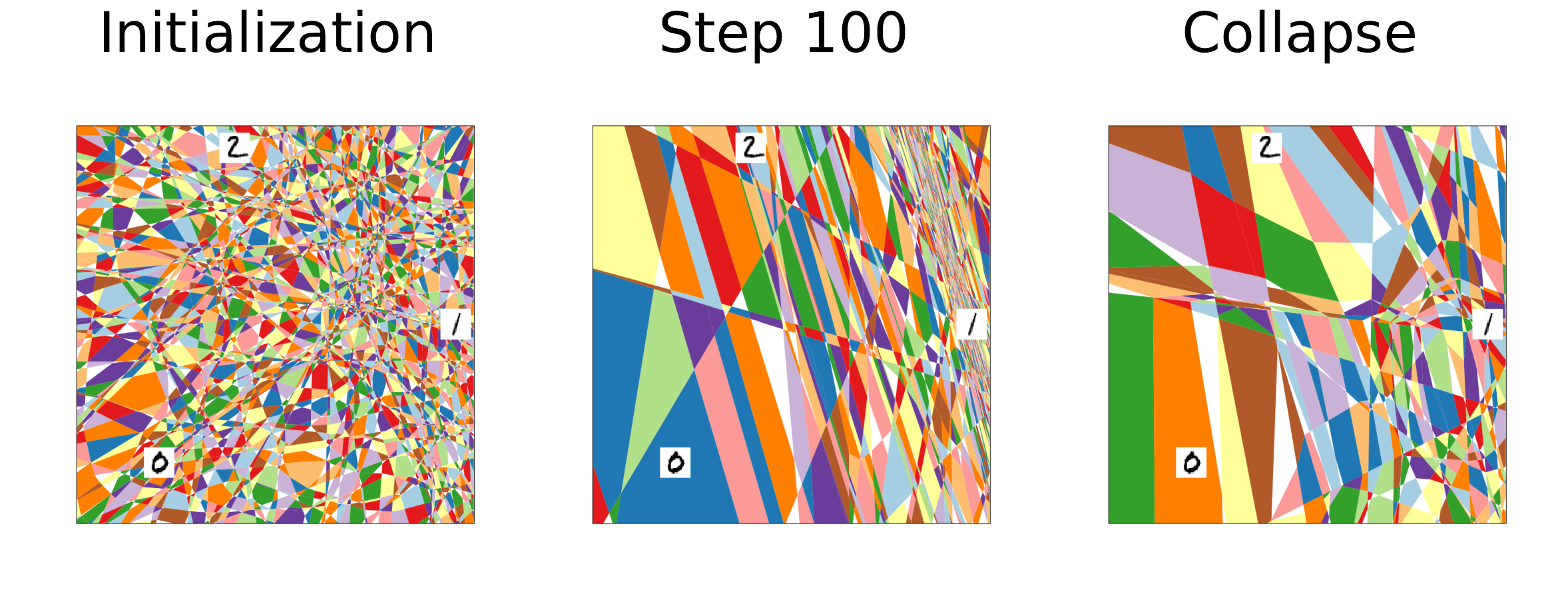}}

   \caption{Illustration of region evolution during representation collapse. The network’s activation space transitions from a highly partitioned structure at initialisation to regions with high volume by step 100, where most linear regions have merged. This trajectory only worsens until collapse.}
   \label{fig:region_collapse}
\end{figure}

We show that the polytopal space can indicate representation collapse early within training. To simulate collapse, we remove the prediction head from the SimSiam framework. \Cref{fig:num_regions_rep_std} jointly tracks the number of linear regions and the standard deviation of feature representations over 700 steps. The representation standard deviation reaches its highest value of 0.02 at 100 steps, then fluctuates for the next 300 steps. The standard deviation only starts to gradually decrease at the 400th step. 
In contrast, the number of regions rapidly decreases by the 100th step. By approximately 150 steps, the network retains only about 5\% of the regions present at initialisation. After this point, the number of active regions stabilises between 5\% and 20\% of the original count. \Cref{fig:region_collapse} depicts the partitioned input space at initialization, 100th step and at the end of training. The network transitions from a highly partitioned structure at initialisation to progressively larger regions at the 100th step, until it collapses.

\subsection{Discussion}

The analysis of the results reveals a clear distinction in the regional evolution between supervised and SSL methods. SSL exhibits restrained region growth while supervised methods continuously partition its input space. As a result, SSL methods require substantially fewer regions to achieve comparable accuracy performance. The reduced region count in SSL corresponds to larger average region volumes. Specifically, contrastive SSL methods exhibit lower eccentricity compared to self-distillation methods. The contrastive objective produces a greater number of regions than self-distillation methods and is more isotropic. 
Lastly, a comparison of the representation standard deviation and the number of regions across training steps indicates that collapse becomes evident earlier in the geometric space. This is a result of regions merging rapidly with no repulsive force in the representation space to discourage shared activation patterns.

\section{Conclusion}

%-----RESTATING THE AIMS OF THE STUDY-----
The study investigates the local distribution of linear regions produced by Self-supervised learning methods (SSL). Furthermore, we compared how supervised, contrastive, and self-distillation methods progressively learn within the geometric space.
%-----SUMMARIZING MAIN RESEARCH FINDINGS-----
Our findings reveal that the geometric space remains a promising approach for assessing representation quality.
% Supervised methods achieve high performance by continuously partitioning the input space. Self-supervised methods rapidly decrease the number of regions early in training, then steadily increase to help fit the data distribution. Contrastive SSL, on average, produce more regions than self-distillation methods by leveraging the repulsive force offered by negative samples. Finally, the geometry of the polytopal space provides an early indicator of representation collapse. 
%-----PROVIDING RECOMMENDATIONS FOR FUTURE RESEARCH-----
Future research could investigate techniques to directly optimise the polytopal space. A novel regularisation strategy could encourage fewer regions with greater volume and controlled eccentricity to improve the robustness and transferability of the model. Out-of-Distribution detection benefits from identifying the location of in-distribution data in the representation space. In the geometric space, a highly partitioned input space may indicate overfitting, revealing regions densely aligned with in-distribution data. Lastly, in continual learning, researchers could leverage self-distillation methods to create as few regions as possible while avoiding collapse within the current task. When new tasks are introduced, contrastive self-supervised methods can then expand the polytopal space by generating additional regions to accommodate new data.
%-----add a short discussion on how the learning of this paper could contribute to improvements-----

\section{Acknowledgment}

We extend our gratitude for the support given by the Center for Artificial Intelligence (CAIR), Institute for Intelligent Systems and the University of Johannesburg, South Africa.
{
    \small
    \bibliographystyle{ieeenat_fullname}
    \bibliography{main}
}

\clearpage
\setcounter{page}{1}
\maketitlesupplementary

\section*{Introduction}

This supplementary section provides additional details about our experimental setup. We outline the full hyperparameter configurations used for each model across both benchmark datasets. We further elaborate on the high-dimensional projection strategy employed. We also expand on the high-dimensional projection strategy adopted. Finally, we include additional illustrations of the results from the FashionMNIST dataset and include supplementary visualisations of the geometric structure of linear regions at the final training epoch for all models.

\section*{Hyperparameters}

\begin{table}[htbp]
\caption{Training hyperparameter configurations for each model on MNIST dataset.}
\begin{center}
\resizebox{\columnwidth}{!}{%
\begin{tabular}{lrrrrrrrrrr}
\toprule
Models  \\
& Epochs & Batch Size & Learning Rate & Weight Decay & Temperature \\
\toprule
\; Classifier                  & \multirow{8}{*}{\rot{100 for each}} & 256 & 0.01 & 0.0 & - \\
\; Triplet loss                && 256 & 0.02 & \multirow{6}{*}{\rot{1.0e-5 for each}} & 0.05 \\
\; SupCon                      && 256 & 0.01 && 0.05 \\
\; SimCLR                      && 256 & 0.01 && 0.2 \\
\; MoCo                        && 64 & 0.008 && 0.2 \\
\; Simsiam                     && 256 & 0.04 && - \\
\; BYOL                        && 256 & 0.04 && - \\
\bottomrule
\end{tabular}
}
\end{center}
\end{table}

\begin{table}[htbp]
\caption{Training hyperparameter configurations for each model on FashionMNIST dataset.}
\begin{center}
\resizebox{\columnwidth}{!}{%
\begin{tabular}{lrrrrrrrrrr}
\toprule
Models  \\
& Epochs & Batch Size & Learning Rate & Weight Decay & Temperature \\
\toprule
\; Classifier                  & \multirow{8}{*}{\rot{100 for each}} & 256 & 0.02 & 0.0 & - \\
\; Triplet loss                && 256 & 0.04 & \multirow{6}{*}{\rot{1.0e-5 for each}} & 0.05 \\
\; SupCon                      && 256 & 0.01 && 0.05 \\
\; SimCLR                      && 256 & 0.01 && 0.2 \\
\; MoCo                        && 64 & 0.01 && 0.2 \\
\; Simsiam                     && 256 & 0.04 && - \\
\; BYOL                        && 256 & 0.04 && - \\
\bottomrule
\end{tabular}
}
\end{center}
\end{table}

\begin{figure}[t]
  \centering
  \centerline{\includegraphics[width=\columnwidth]{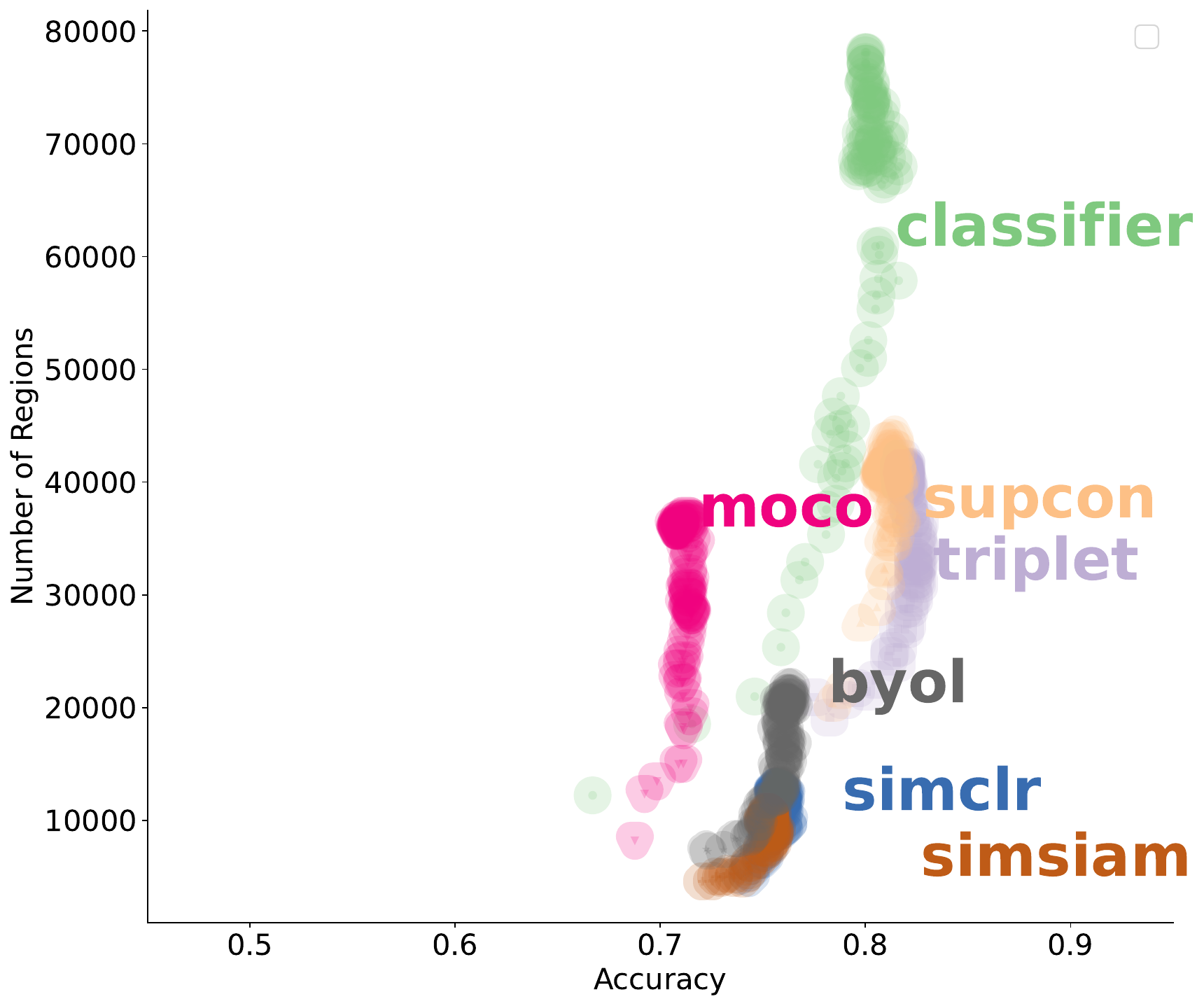}}

   \caption{Number of regions and accuracy achieved by supervised and self-supervised methods on FashionMNIST dataset. The opacity reflects the progression of training over epochs. Low-opacity points depict results achieved early in training, while darker points represent results achieved later.}
\end{figure}

\section*{High-dimensional Projection}

We describe the process used to project a two-dimensional point to a high-dimensional space~\cite{novak2018sensitivity, rolnick2017power}. Let $x_0, x_1 \text{ and } x_2 \in \mathbb{R}^d$ be three sampled input images. We construct:
\begin{enumerate}
    \item the circumcenter of the triangle produced by the three points,
    \item the orthonormal basis vector of the plane,
    \item an affine mapping that transforms high-dimensional coordinates into the two-dimensional input space.
\end{enumerate}
Let $\mathbf{v}_1 = x_1 - x_0$ and $\mathbf{v}_2 = x_2 - x_0$. The circumcenter $C$ of the triangle formed by $x_0, x_1 \text{ and } x_2$ can be expressed as:
\begin{equation}
    C = x_0 + \beta_{0} \mathbf{v}_1 + \beta_{1} \mathbf{v}_2
\end{equation}

\begin{figure*}[t]
  \centering
  \centerline{\includegraphics[width=0.9\textwidth]{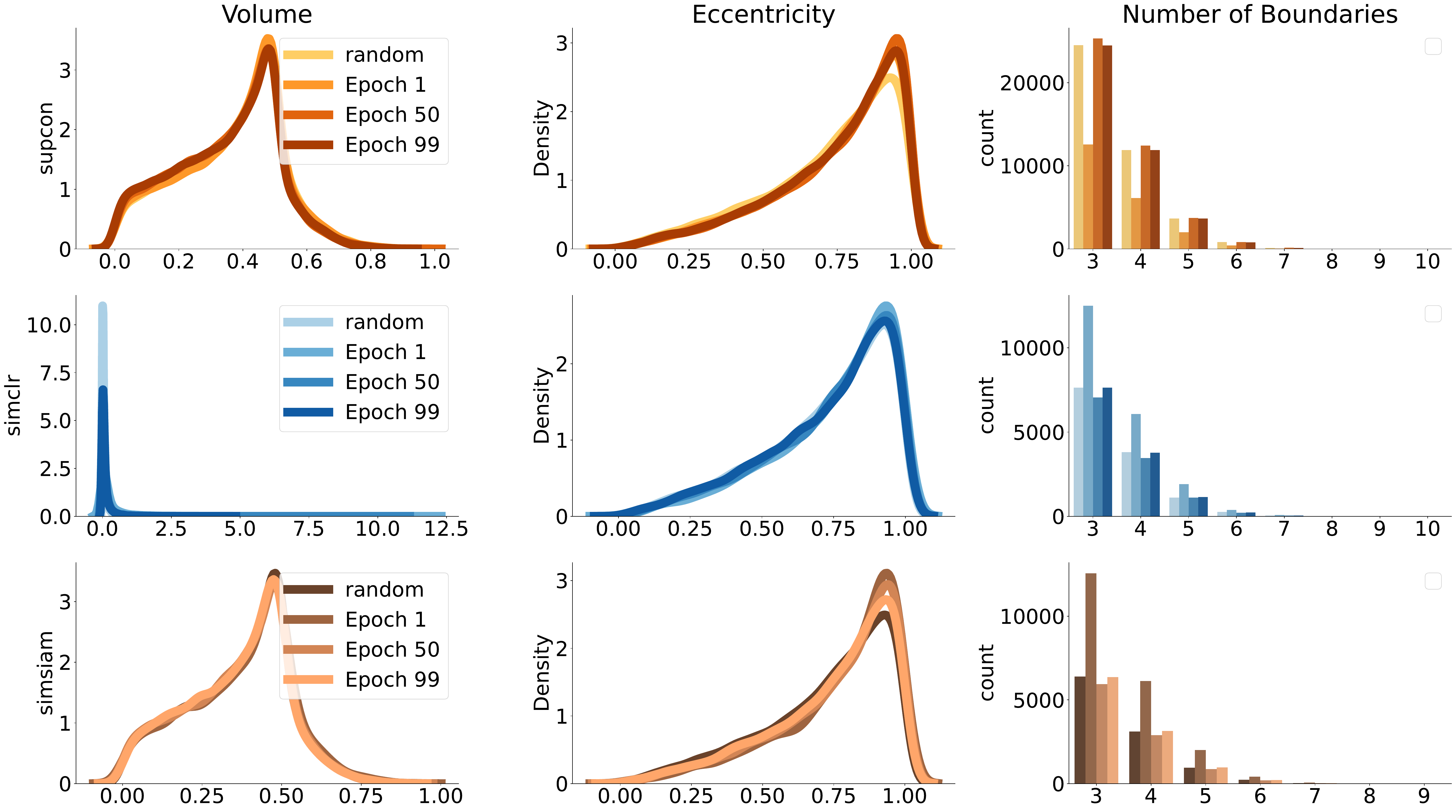}}

   \caption{Evolution of region area, eccentricity, and boundary count distributions across epochs for SupCon, SimCLR, and SimSiam on FashionMNIST dataset.}
\end{figure*}

where $\beta_{0}, \beta_{1} \in \mathbb{R}$ are the coefficients that satisfy the linear system
\begin{equation}
    M \begin{bmatrix} a \\ b \end{bmatrix} = \frac{1}{2} b
\end{equation}
such that
\begin{equation}
    M = \begin{bmatrix} \mathbf{v}_1^T \mathbf{v}_1 & \mathbf{v}_1^T \mathbf{v}_2 \\ \mathbf{v}_1^T \mathbf{v}_2 & \mathbf{v}_2^T \mathbf{v}_2 \end{bmatrix}, \ b = \frac{1}{2} \begin{bmatrix} \Vert \mathbf{v}_1 \Vert^2 \\ \Vert \mathbf{v}_2 \Vert^2 \end{bmatrix}
\end{equation}
solving the linear system gives the circumcenter.
Next we use the Gram–Schmidt process to obtain the orthonormal basis vectors. Given the center vectors $w_1 = x_1-C \text{ and } w_2 = x_2 - C$. If $u_1 = \frac{w_1}{\Vert w_1 \Vert}$ is the normalized vector then:
\begin{equation}
    w_2^{\perp} = w_2 - \bigg( \frac{w_2^{\top} u_1}{u_1^{\top} u^1} \bigg) u_1
\end{equation}
points in the perpendicular direction. Let $u_2 = \frac{w_2^{\perp}}{w_2^{\perp}}$ then the orthonomal basis vector is defined as $\{ u_1, u_2 \} \subseteq \mathbb{R}^d$.
The input $x$ can be obtained from the point $ [ a, b ]$ in the two-dimensional space by setting
\begin{equation}
    x(a, b) = C + a u_1 +b u_2
\end{equation}
similarly
\begin{equation}
    x = \begin{bmatrix} u_1 & u_2 & C \end{bmatrix} \begin{bmatrix} a \\ b \\ 1 \end{bmatrix}
\end{equation}
such that $T = \begin{bmatrix} u_1 & u_2 & C \end{bmatrix}$ is the transformation matrix from a two-dimensional space to a high-dimensional space.

\begin{figure*}[t]
  \centering
  \centerline{\includegraphics[width=\textwidth]{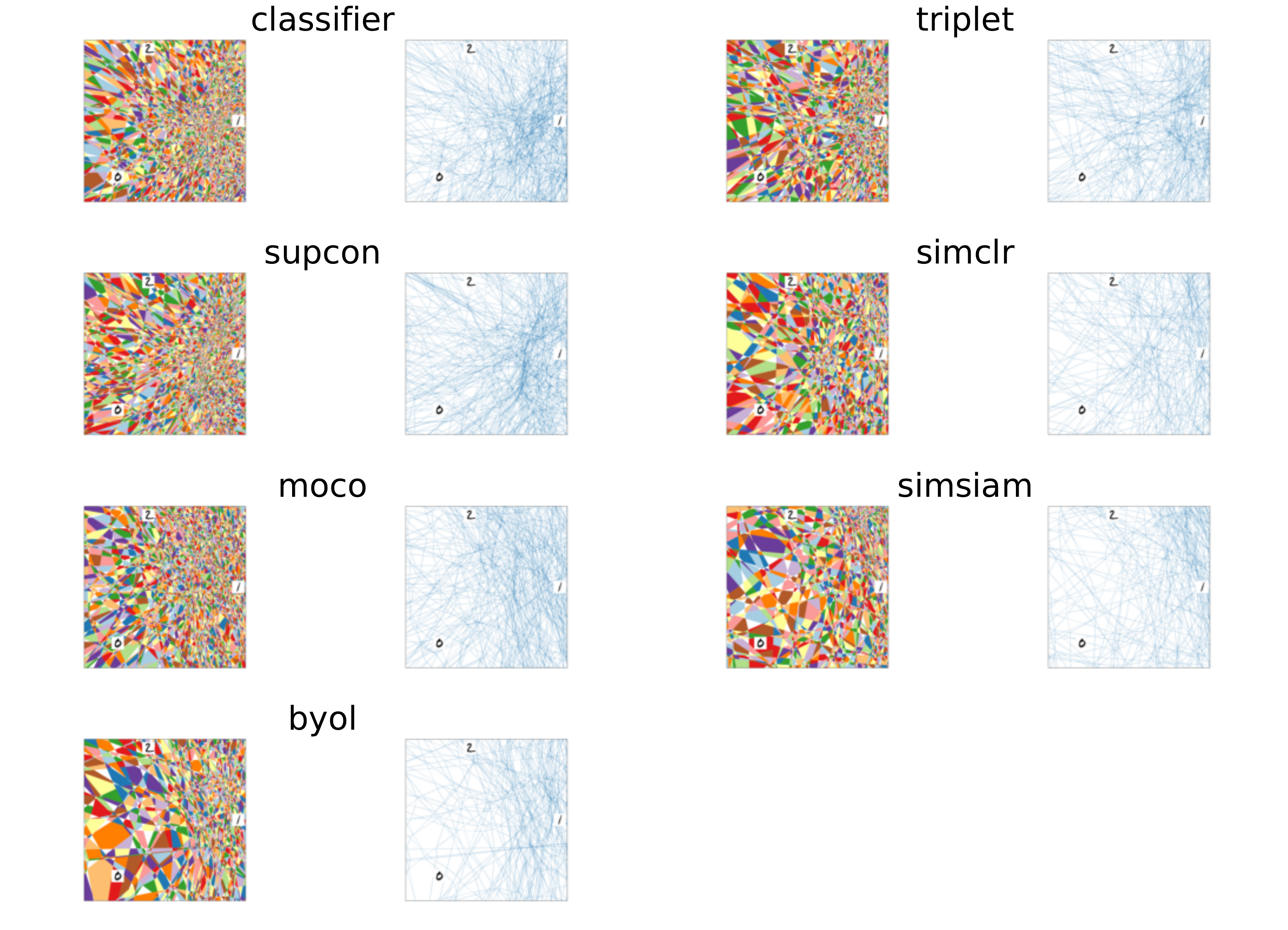}}

   \caption{Final geometry of regions at the 100th epoch for each model on the MNIST dataset. Supervised methods (Classifier, Triplet and SupCon) produce highly dense and smaller regions. Self-supervised methods (SimCLR, MoCo, SimSiam, and BYOL) produce less dense regions with larger volumes. Contrastive self-supervised methods (MoCo and SimCLR) have linear regions that are more dense than the self-distillation methods (Simsiam and BYOL). The use of negatives as a repulsive force within the embedding space partitions the geometric space further.}
\end{figure*}

\begin{figure*}[t]
  \centering
  \centerline{\includegraphics[width=\textwidth]{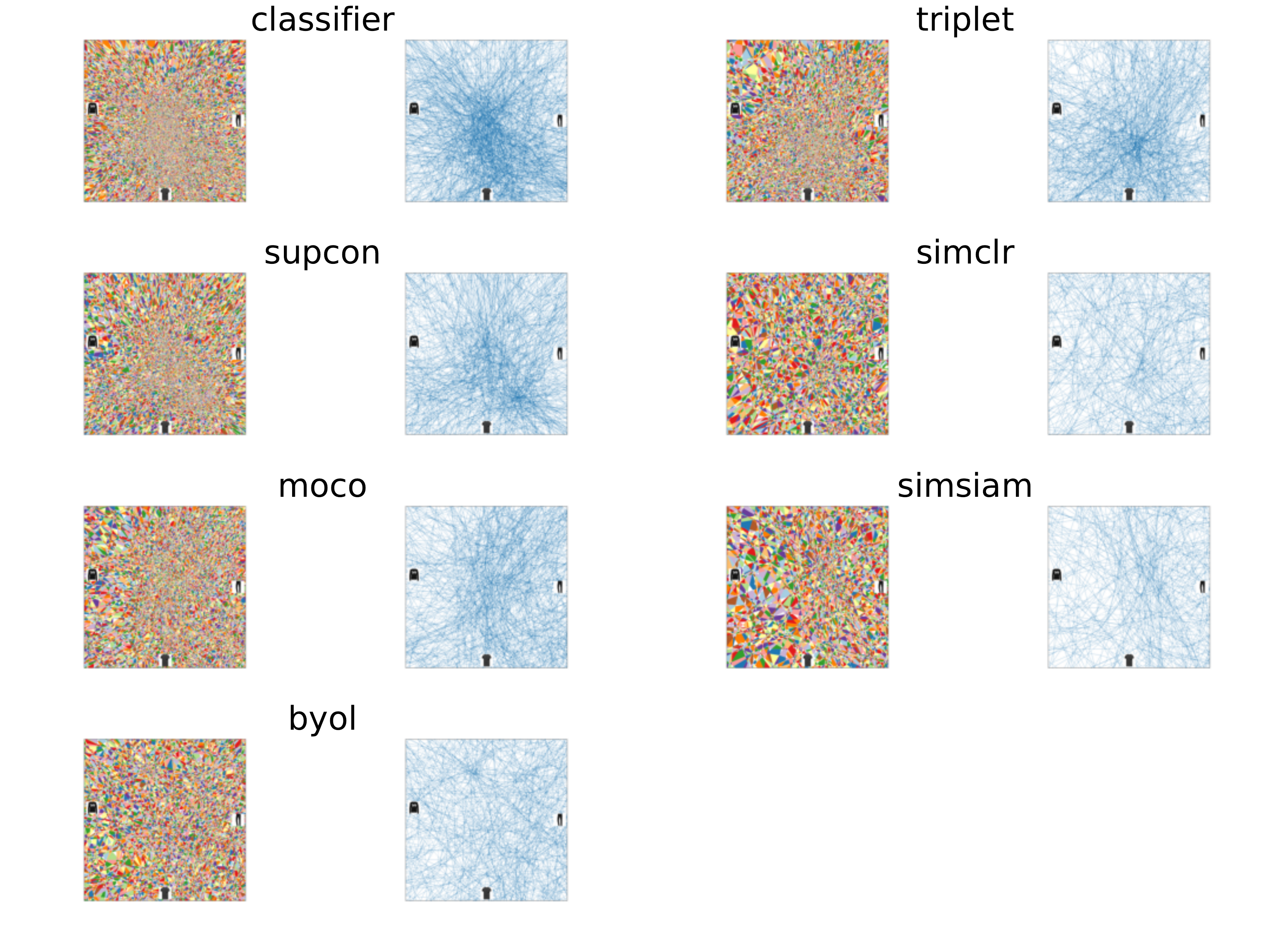}}

   \caption{Final geometry of regions at the 100th epoch for each model on the FashionMNIST dataset. The same qualitative trends observed on MNIST are reproduced here. Regions consolidate in self-distillation methods and finer partitioning in supervised and self-supervised contrastive models.}
\end{figure*}

\end{document}